# Using Deep Learning to Explore Local Physical Similarity for Global-scale Bridging in Thermal-hydraulic Simulation

Han Bao[1], Nam Dinh[2], Linyu Lin[2], Robert Youngblood[1], Jeffrey Lane[3], Hongbin Zhang[1]

[1]*Idaho National Laboratory, Idaho Falls, ID 83415, USA*
[2]*North Carolina State University, Raleigh, NC 27695, USA*
[3]*Zachry Nuclear Engineering Inc., Cary, NC 27518, USA*

**ABSTRACT**

In the model development and assessment process, data is usually collected from reduced-scale facilities, and a scaling analysis is needed to ensure the data applicability with respect to the prototypic systems. During a classical analysis, some issues, including scaling distortion and scaling uncertainty, have become stumbling blocks for ensuring the validity of the modeling and simulation tool and for demonstrating the safety of a nuclear-reactor design. As a result, current system thermal-hydraulic codes have limited credibility in simulating real plant conditions, especially when the geometry and boundary conditions are extrapolated beyond the range of test facilities. Although some advanced coarse-mesh codes have been widely used in system-level safety analyses due to their balance on computational efficiency and simulation accuracy, the verification and validation (V&V) of these codes still suffers from a lack of prototypic validation data. However, considering that mesh size is one of the model parameters for these coarse-mesh codes with simplified boundary-layer treatment, the mesh-induced error and model error are tightly connected, which makes it difficult to evaluate mesh effect or model scalability independently as used in classical scaling analysis. This paper proposes a data-driven approach, Feature Similarity Measurement (FSM), to establish a technical basis to overcome these difficulties by exploring local patterns using machine learning. The underlying local patterns in multiscale data are represented by a set of physical features that embody the information from a physical system of interest, empirical correlations, and the effect of mesh size. It is argued that the interpolation in these local physical features can reflect the extrapolation of global conditions. After performing a limited number of high-fidelity numerical simulations and a sufficient amount of fast-running coarse-mesh simulations, an error database is built, and deep learning is applied to construct and explore the relationship between the local physical features and simulation errors. As a result, a data-driven model can be developed to provide an accurate estimate on the simulation error even when global scale gaps exist. Case studies based on mixed convection have been designed for demonstrating the capability of data-driven models in bridging global scale gaps. Results show that the predictions by well-trained data-driven models are more accurate with higher similarity between training and target data.

## 1. INTRODUCTION

Scaling is the process of assessing the similarity of phenomena that occurred and was observed in a reduced-scale test facility and the full-scale nuclear-reactor-plant (NPP) application. Because it is impractical to perform experiments under prototypic conditions, the prediction of prototype-scale processes is normally made by models developed based on scaled experiments. These scaled experiments are designed by decomposing and down-scaling the full-scale applications into a series of tests that attempt to isolate individual phenomena. These are often referred to as separate-effects tests. The phrase "scaling issues" refers to difficulties and complexities stemming from the applicability of the data measured in the scaled experiments to the conditions expected in the prototype. These issues arise from the impossibility of obtaining transient data from the prototype system under off-nominal conditions. Solving a scaling issue implies developing approaches, procedures, and data suitable for predicting the prototype's performance utilizing scaled models or data (Bestion et al., 2016).

Scale-invariant approaches are the ideal approach to explore and predict behaviors in real full-scale applications. Scale invariance represents entities that are independent of scale, such as physics and direct numerical simulation (DNS). There are two kinds of scale-invariant approaches: (1) a full-scale (or physics-conserved) experiment, which is (presumably) independent of facility scale or (2) DNS modeling, in which local information is solved accurately with very fine mesh. However, full-scale experiments are hard to build while many full-scale tests are required. Meanwhile DNS is a computationally expensive way to deal with the system scenario simulations. Reduced-order models—e.g., large-eddy simulation (LES), Reynolds-averaged Navior-Stokes (RANS) models, and system codes are not scale-invariant approaches. That is where scaling distortion, which refers to any discrepancy between the applied scaled parameter and the referenced plant parameters, exists in system simulation. Although scaling distortion exists, advanced coarse-mesh computer codes are widely used in NPP safety analyses because of both cost and time effectiveness. The effect of scaling on the model error and consequent uncertainty calibrated using data from scaled experiments greatly influences the accuracy of simulation and leads to an unknown error and uncertainty. The uncertainty due to scaling effect is called scaling uncertainty. Although several scaling methods have been developed to quantify the distortions—hierarchical two-tier scaling (H2TS, see (Zuber et al., 1998), fractional scaling analysis (FSA, see (Zuber et al., 2007) and dynamical systems scaling (Reyes, Frepoli & Yurko, 2015)—it is recognized that a complete similarity between the prototype and the model cannot be achieved, particularly in a complex NPP system.

In addition to scaling uncertainty, there are other errors and sources of uncertainty affecting the accuracy of system thermal-hydraulic analysis. Typically, there are two kinds of computational codes used for system thermal-hydraulic analysis: system codes (e.g., RELAP, TRACE) that describe the reactor system as a network of simple control volumes connected with junctions and computational fluid dynamics (CFD)-like codes (e.g., GOTHIC) that provide a three-dimensional (3D) simulation capability using coarse-mesh configurations with the sub-grid phenomena in boundary layer that is well captured by adequate constitutive correlations (e.g., wall functions and turbulence models). Compared with standard system codes (with much loss of local information) and standard fine-mesh CFD codes (with huge computational cost), these coarse-mesh, CFD-like codes have natural advantages and have been widely used to achieve sufficient accuracy for long-term thermal-hydraulic simulation of multicomponent system (Chen et al., 2011, Ozdemir, George & Marshall, 2015, Bao et al., 2016, Bao et al., 2018b). They solve the conservation equations for mass, momentum, and energy for multicomponent multiphase flow. One of the main error sources using these codes is model error due to physical simplification and mathematical approximation of the applied models, correlations, and assumptions. For thermal-hydraulic simulations using these codes, the key local phenomena in the near-wall region are friction, turbulence, and heat transfer. Respective models or correlations are applied for the simulation where characteristic lengths are introduced as one of the key parameters. The calculation of characteristic length is, by default, executed using local mesh size. Local mesh size is treated as one of the model parameters that determine whether the correlations are applied in their applicable ranges. Another main error source, mesh-induced error, is also related to local mesh size, and is defined as the information loss of conservative and constitutive equations by applying time- and space-averaging approaches.

Traditional V&V frameworks analyze model and numerical errors separately. However, these methodologies are impractical in these coarse-mesh CFD-like codes because mesh size is treated as a model parameter and mesh convergence is not expected. Usage of these coarse-mesh codes for the evaluation of safety margins, the training of reactor operators, the optimization of the plant design, and the development of emergency operating procedures is wide and has been growing. At the same time, V&V has been recognized as a mandatory prerequisite of their applications. In 1974, the United States Nuclear Regulatory Commission (USNRC) published rules for loss of coolant accident (LOCA) analysis in 10 CFR 50.46 (USNRC, 1974a) and Appendix K (USNRC, 1974b). It established initial licensing procedures with a conservative approach. However, considering the issue of scaling distortion, conservatism proved in reduced-scale tests may become invalid in the full-scale plants. As a result, the USNRC initiated an effort to develop and demonstrate the best-estimate plus uncertainty (BEPU) method. This provides nuclear plant operators with more economic gains and less conservatism. The code scaling and applicability uncertainty (CSAU) method was formulated to provide more-realistic estimates on plant safety margins for a large-break LOCA in a pressurized water reactor (PWR) in 1990 (Boyack et al., 1990). In 2005, USNRC issued another important contribution, Regulatory Guide (RG) 1.203, "Evaluation Model Development and Assessment Process (EMDAP)," as a demonstration of a validation framework that is considered acceptable for the best-estimate calculations of NPP transient and accident analysis (USNRC, 2005). Mesh effect on code and model scalability was not fully considered in CSAU and EMDAP. They both recognized the existence of mesh sensitivity and required a mesh-sensitivity study to make sure important figures of merit (e.g., peak cladding temperature) were not significantly impacted. They assumed that a "constant" uncertainty was introduced between the scaled tests and plant application by applying the same modeling guidelines and consistent mesh configurations. However, the fact that mesh size could be one of the key model parameters and influence code and model applicability was not fully considered because mesh sensitivity was performed before the code-scalability analysis. Although the importance of code-scalability analysis has been well recognized and emphasized in both CSAU and EMDAP, it is difficult to perform with the tight connection between main error sources and error or uncertainty propagation due to scaling uncertainty. Thus, a new approach that comprehensively assesses the adequacy of evaluation models is needed to bridge scale gap and realize code scalability analysis.

In this paper, a data-driven approach is developed to quantify the uncertainty (simulation error) by exploring local patterns using machine learning. The underlying local patterns in multiscale data are represented by a set of physical features that integrate information from the physical system of interest, empirical correlations, and the effect of mesh size. This approach provides a technical basis for a preliminary development of a data-driven scale-invariant system simulation technology by treating main error sources and scaling uncertainty together. The central idea of this data-driven approach is discussed in Section 2; it is proposed that the similarity of local physics could be identified using deep machine learning to bridge the global scale gap. Machine learning applications in thermal-hydraulic analysis are also reviewed. Section 3 describes the workflow of the proposed approach, scalability of which is investigated based on mixing convection cases studies in Section 4. Section 4 also discusses how to improve machine-learning predictions by considering global conditions in the definitions of local physical features. Conclusions and future work are summarized in Section 5.

## 2. TECHNICAL BACKGROUND

### 2.1. Concept of Physics Coverage Condition

Over the past few decades, many designs of nuclear reactor, with different systems, geometries and powers, have been proposed. The respective global physical conditions might be an "extrapolation" to previous simulations and experiments, which brings large uncertainty to the demonstration studies using previously developed models. Relevant thermal-hydraulic experiments with a wide range of scale must be redesigned to validate the applicability of codes and models for these new global conditions. However, in some respects, the local physics such as the interaction between liquid, vapor, and heat structure are independent of the global physics. In other words, it is possible that local physical parameters or variables in the local cells are similar even when the global physical condition totally changes.

Global and local physics are differentiated: the former indicates a global or macroscopic state, observation and deduction of the physical condition, such as the dimension, geometry, structure, boundary condition and non-dimensional parameters that represent the underlying global physics. The latter refers to the microscopic state, observation and deduction of the physical condition. For example, the global physics of turbulent flow can be characterized using Reynold (Re) number. No matter how the Re number or geometry changes, the local physics always involve turbulence if the Re number is big enough. According to the identification of global physics and local physics, four different physics coverage conditions (PCCs) are classified: global interpolation through local interpolation (GILI), global interpolation through local extrapolation (GILE), global extrapolation through local interpolation (GELI) and global extrapolation through local extrapolation (GELE).

For instance, there are several cases for single-phase fully developed flow in a pipe of diameter D: local physical conditions and values of $Re_D$ representing the global physical conditions are listed in Table 1. Assuming some cases as existing data and others as target simulations, four different physics coverage conditions can be specified as shown in Figure 1.

- *GILI condition* represents the situation where both global and local physical conditions of the target case (Case 4) are identified as an interpolation of existing cases (Cases 3 and 5). The physics of the target case are globally and locally "covered" by existing cases. The model developed using the data from Case 3 and 5 is reliable to predict Case 4.
- For the *GILE condition,* even if global physical condition of the target case (Case 2) is covered by existing cases (Case 1 and 3), data from existing cases are not able to predict the target case because local physics are different. In fact, the models developed from experiments of laminar flow or turbulent flow are not applicable for transition prediction.
- *GELE condition* has the same problem: models developed from the experiments of laminar flow are not applicable for turbulence prediction. In GILE and GELE condition, existing data do not contain instructive information for the target case, so it is useless no matter how much data are used.
- *GELI condition* indicates a situation in which global physical condition of the target case (Case 5) is identified as an extrapolation of existing cases (Case 3 and 4), but local physics are similar, as turbulent flow. The values of some representative parameters (e.g., local velocity gradients) are even interpolative in the existing cases. Unlike in GILE or GELE conditions, the local similarity in GELI condition provides feasibility to take benefits from existing data to estimate the target case.

Table 1. Example of Different Global and Local Physical Conditions

| Case | Global Physical Condition | Local Physics |
|---|---|---|
| 1 | $Re_D = 10^2$ | Laminar Flow |
| 2 | $Re_D = 2.5 \times 10^3$ | Laminar–Turbulent Transition |
| 3 | $Re_D = 1 \times 10^4$ | Turbulent Flow |
| 4 | $Re_D = 2 \times 10^4$ | |
| 5 | $Re_D = 3 \times 10^4$ | |

Instead of endlessly evaluating applicable ranges of models and scaling uncertainty, exploring the similarity of local physics opens another door to overcome the globally extrapolative problems. For the specific physics of interest, i.e., the specific physical models, local mesh sizes and numerical solvers are treated as an integrated model. Data obtained from this integrated model can be used to construct a library that identifies and stores the local similarities in different global physical conditions. This library is improvable as new qualified data are available. Once the library is built, machine-learning algorithms are applied to find patterns in the local data and to make predictions.

| **GILE** | **GELE** |
|---|---|
| Global Interpolation through Local Extrapolation Case: 1 + 3 → 2 | Global Extrapolation through Local Extrapolation Case: 1 + 2 → 3 |
| Global Interpolation through Local Interpolation Case: 3 + 5 → 4 | Global Extrapolation through Local Interpolation Case: 3 + 4 → 5 |
| **GILI** | **GELI** |

**Figure 1. Illustration of Physics Coverage Condition Considering Global and Local Physics**

Currently, the grounded-physics coverage condition for analysis and code/model V&V is the GILI condition in which existing data or experience has the capability to estimate the target case due to both the global and local similarities. GELI condition has this potential capability when sufficient data, appropriate physical features and machine-learning algorithm are available. The extrapolation of global physics indicates different global physical conditions, such as a set of characteristic non-dimensional parameters, or different IC/BCs (Initial Conditions / Boundary Conditions), or different geometries/structures, or dimensions. The interpolation of local physics implies two definitions: (1) from the perspective of traditional phenomena decomposition, the existing cases and target case are designed for the local physics with similar length and time scales, such as the turbulence example discussed above and (2) from the perspective of data characteristics, the underlying local physics of these cases is assumed to be represented by a set of physical features (PFs), and a major part of the data of the target case is covered or similar to the data of existing cases. This similarity is dependent on the identification of physical features and on data quality and quantity.

Therefore, the proposed data-driven approach, feature-similarity measurement (FSM), aims to identify the local physical features, measure the data similarity of defined physical features, and investigate the relationship between physical-feature similarity and accuracy of machine-learning prediction in the GELI condition. Some previous efforts have been made based on the FSM approach. The predictive capability of FSM for globally extrapolative conditions has been preliminarily evaluated by a turbulent-mixing case study (Bao et al., 2018a) and a two-phase-flow case study (Bao et al., 2019a). A data-driven framework (optimal mesh/model information system, OMIS) was developed and demonstrated to improve applications of the coarse-mesh codes by predicting their simulation errors and suggesting the optimal mesh size and closure models based on the FSM approach (Bao et al., 2019b). This physics-guided data-driven approach is also applied to realize the computationally efficient and scalable CFD prediction of bubbly flow(Bao, Feng et al., 2019). The basic idea of FSM was illustrated in Figure 2. Based on the involved phenomena, closure models and mesh configuration, local physical features are defined as inputs of surrogate models while the outputs are local simulation errors.

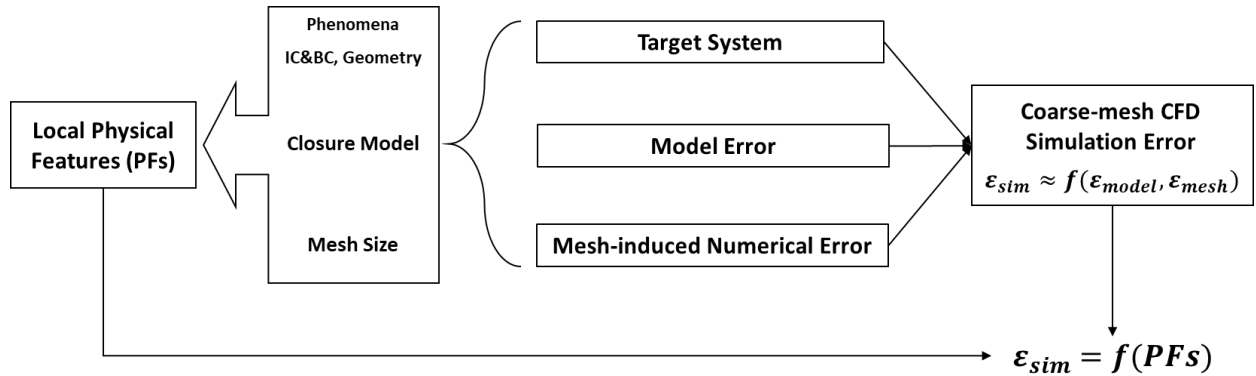

Figure 2. Basic Idea of FSM: Surrogate Modeling of Simulation Error and Physical Features (Bao, Feng et al., 2019)

## 2.2. Literature Review: Data-driven Modeling Application on Fluid Dynamics

Recently, the rise of performance computing in fluid dynamics has led to large high-fidelity data generation from DNS and well-resolved LES for the training and development of data-driven turbulence closures. (Tracey, Duraisamy & Alonso, 2015) used feedforward neural networks (FNNs) to predict the Reynolds stress anisotropy and source terms for turbulence-transport equations. Ling explored the capability of random forests (RFs) and FNNs in learning the invariance properties and concluded that RFs are limited in their ability to predict the full anisotropy tensor because they cannot easily enforce Galilean invariance for a tensor quantity (Ling, Jones & Templeton, 2016). Wang proposed to apply RFs to predict the Reynolds stress discrepancy instead of directly predicting Reynolds stress (Wang, Wu & Xiao, 2017). Recently, Zhu developed a data-driven approach to predict turbulence Reynolds stress in the RANS model by leveraging the potential of massive DNS data (Zhu, Dinh, 2017). These approaches focused on how to improve RANS turbulence modeling without considering numerical errors. Similarly, a data-driven approach was developed to estimate the model error from boiling closures (Liu et al., 2018). Hanna proposed a coarse-grid (CG)-CFD approach using machine learning algorithms to predict the local errors (Hanna et al., 2020). This work aims at correction of mesh-induced error in CG-CFD without considering the model errors that may be introduced in CFD applications on thermal-hydraulic analysis.

All these data-driven approaches reviewed above are not designed for CFD-like or coarse-mesh CFD codes. These efforts analyzed model error and mesh-induced error separately, with another fixed; this is inapplicable to the coarse-mesh methods where mesh size is treated as a model parameter and mesh convergence is not expected. The uncertainty propagation due to scaling uncertainty makes it more difficult to estimate the simulation error when using these codes for system analysis. The FSM approach is developed to deal with these two error sources together. The machine-learning model trained in the FSM approach treats physical correlations, coarse mesh sizes, and numerical solvers as an integrated model that can be considered as a surrogate of governing equations and closure correlations of low-fidelity code. Low-fidelity simulation refers to the simulations with coarse-mesh configurations and not fully validated closure models. The development of this integrated model does not need relevant prior knowledge, and purely depends on data. Compared with these data-driven efforts on adiabatic flow problems, the FSM approach is successfully applied to thermal-hydraulic modeling and simulations, which are described in the following case studies. Deep FNNs (DFNN) with Bayesian regularization algorithm for backward propagation (MacKay, 1991) are applied as the machine-learning tool in this work.

## 3. DESCRIPTION OF THE PROPOSED DATA-DRIVEN APPROACH

Consider a physical system that is governed by a set of non-linear equations. The physical system can be simulated using a coarse-mesh code as,

$$F_{LF}(\vec{V_{LF}}(\vec{x},t), \lambda_{LF}, \delta_{LF}) = 0 \tag{1}$$

where $F_{LF}$ is the set of governing equations and constitutive equations as a low-fidelity model, $\vec{V_{LF}}$, $\lambda_{LF}$ and $\delta_{LF}$ represent the model variables, the model information (model forms and relative parameters), and the coarse mesh size used in the low-fidelity simulation.

Simulation error ($\varepsilon$), including model error, mesh error, and other numerical errors exists, even if the best possible set of parameters, models and mesh sizes have been inferred. Given the true solution as $\vec{R_T}$ for the same physical condition, then the output quantities of interest can be expressed as,

$$\vec{R_T} = \vec{R_{LF}}(\vec{V_{LF}}, \lambda_{LF}, \delta_{LF}) + \varepsilon + \epsilon \tag{2}$$

where $\vec{R_{LF}}$ represents the output of the low-fidelity simulation, and $\epsilon$ is the measurement error. Then $\varepsilon$ is expressed as below for a given physical condition.

$$\varepsilon = (\vec{R_T} - \epsilon) - \vec{R_L}(\vec{V_{LF}}, \lambda_{LF}, \delta_{LF}) \tag{3}$$

If the measurement error is considered to be negligible, one can conclude that the low-fidelity simulation error $\varepsilon$ is determined by physical condition represented by $\vec{R_T}$ and $\vec{V_{LF}}$, model information (model form and parameter, $\lambda_{LF}$) and the coarse mesh size $\delta_{LF}$ used for the low-fidelity simulation. However, the function form shown in Equation (3) is difficult to identify. By integrating three factors together, the FSM approach identifies a group of physical features that represent the local patterns of the physics of interest, and then utilizes machine-learning techniques to specify the relationship between these local physical feature groups and local simulation errors of quantities of interest (QoIs). The basic assumptions for the application of the FSM approach are (1) length scale of the physics of interest is large enough to be captured by coarse-mesh modeling and simulation, (2) coarse-mesh codes are able to capture the basic physical behaviors of the system of interest with an acceptable uncertainty to be quantified, (3) simulation error is mainly impacted by model error and mesh-induced error, for which mesh size is one of the key model parameters that makes them tightly connected, and (4) training data are qualified and sufficient for the machine-learning algorithm to learn from and find the intrinsic knowledge of the local physics.

### 3.1. Identification of Local Physical Features

The identification of physical feature s is guided by the physics decomposition and model evaluation. To take physics scalability and regional information into consideration, the physical-feature group should include the gradients of local variables and the local physical parameters that are able to represent local physical behaviors applied in crucial closure relationships, as shown in Figure 3. This ensures that the physical information of the physical system, model information, and the effect of mesh size are integrated and well represented in the physical-feature group.

The first part of local physical features is gradients of variables, including 1-order and 2-order derivatives of variables calculated using central-difference formulas with boundary information considered. The gradients of local variables imply the regional (or local, surrounding) information that represents the regional physical patterns. More regional information may be involved if higher-order derivatives are added into the local physical-feature group. The second part of the physical-feature group is local parameters that are defined to represent the local physical behaviors or are applied in closure relationships. These parameters representing local physical behaviors are supposed to provide the scalability of physics. Another part of local parameters as physical features is the parameters used or involved in the crucial local-closure correlations for the boundary layer. These parameters contain much information of length scale, model parameter, and wall distance. In the work of OMIS framework (Bao et al., 2019b), all potential physical features are supposed to be identified and ranked according their importance using the RF algorithm.

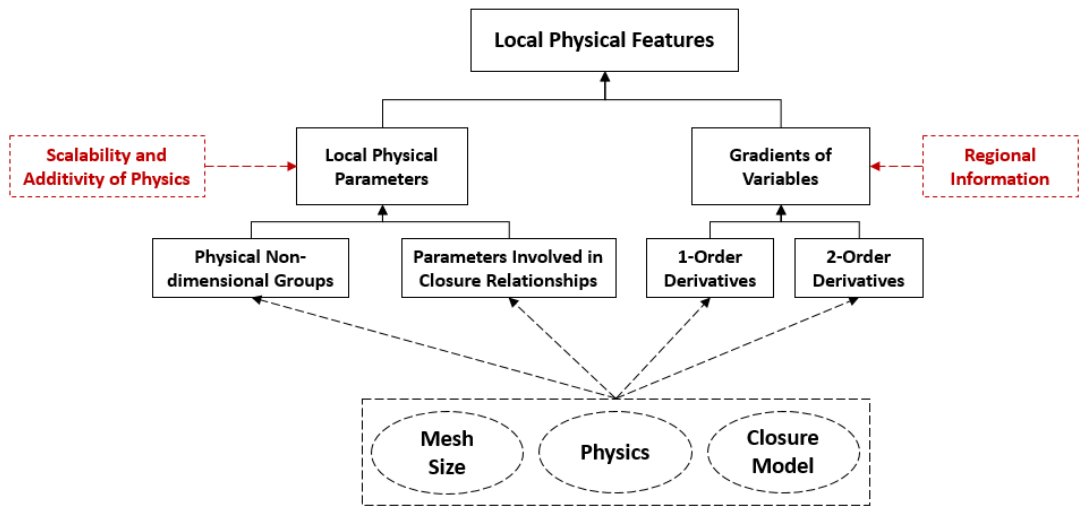

**Figure 3. Identification and classification of Physical Feature (Bao et al., 2019b)**

### 3.2. Measurement of Data Similarity

Several types of distance metrics have been used to measure the distance between a target point and the training dataset. Metrics based on Euclidean distance, such as the nearest-neighbor distance, are easy to compute, but susceptible to noise and memory-consumption because all points in the training dataset are used. Besides, these metrics treat the training data as uncorrelated points and ignore their underlying interactions. Some promising metrics are designed for memory-efficiency by considering the distribution of the training dataset. Mahalanobis distance is defined as the distance between a point and the mean of training data with the covariance matrix. However, the drawback of Mahalanobis distance is its assumption that the training data points yield a multivariate Gaussian distribution. It is inapplicable to deal with the data from thermal-hydraulic simulations, especially for turbulent flows where multimode distributions may be common. Another method, called kernel-density estimation (KDE) is a non-parametric way to estimate the probability density function, which assumes the training-data distribution can be approximated as a sum of multivariate Gaussians. One can use kernel distribution when a parametric distribution cannot properly describe the data or when one wants to avoid making assumptions about the distribution of the data. KDE can be considered as the probability that the point ($\boldsymbol{q}$) locates in the distribution of training data ($\boldsymbol{p}_i, i = 1,2, \ldots, n$). It is expressed as (Scott, 2015),

$$p_{KDE} = \frac{1}{n \cdot h_1 h_2 \ldots h_d} \sum_{i=1}^{n} \prod_{j=1}^{d} k\left(\frac{q_j - p_{i,j}}{h_j}\right) \qquad (4)$$

Where $d$ is the number of variables in $\boldsymbol{q}$ and $\boldsymbol{p_i}$, $k$ is the kernel-smoothing function, and $h_j$ is the bandwidth for each variable. A multivariate kernel distribution is defined by a smoothing function ($k$) and a bandwidth matrix defined by $H = h_1, h_2, \ldots, h_d$, which controls the smoothness of the resulting density curve. Therefore, KDE can be used to measure the distance by estimating the probability of a given point located in a set of training-data points. In this step, the KDE distance is standardized as,

$$d_{KDE} = 1 - \frac{p_{KDE}}{p_{KDE} + 0.1} \qquad (5)$$

Before the calculation of KDE distance, the data of physical feature *s* should be normalized into the range [0, 1]. Then the standardized KDE distance will locate from 0 to 1. Higher value of KDE distance means a higher level of extrapolation. The mean of KDE distance is calculated to represent the averaged distance from the target database to the training database,

$$D_{KDE} = \frac{1}{n} \sum_{i=1}^{n} d_{KDE,i} \qquad (6)$$

t-distributed stochastic neighbor embedding (t-SNE) method (Maaten, Hinton, 2012), a dimensionality-reduction technique, is applied for the visualization of high-dimensional datasets. Relative distances among the points are retained and reflected from high to low dimensionality.

## 4. CASE STUDY ON EXTRAPOLATIONS OF GLOBAL SCALE

In this section, a 2-D cavity with hot air injection on bottom of one sidewall, a vent on the other sidewall and a cold top wall has been modeled to simulate the mixed convection considering turbulence. Three extrapolative situations in GELI condition are designed to investigate the scalability and predictive capability of FSM approach. FSM approach are performed to identify physical features, measure similarity of data, and evaluate the relationship between data similarity and prediction accuracy. Section 4.1 to 4.3 respectively discusses the extrapolation of geometry, boundary condition, and dimension. The physical features are identified as shown in Table 2. Five non-dimensional parameters are defined as: $R$ that includes the turbulent information; $Reynolds$ ($Re$) number that is defined with the consideration of both $Re_f$ in free cells and $Re_W$ in near-wall cells; $Grashof$ ($Gr$) number that approximates the ratio of the buoyancy to viscous force acting on a fluid by considering the local density change which also considers the difference of hydraulic diameter calculations in near-wall cells and free cells; $Richardson$ ($Ri$) number that expresses the ratio of the buoyancy term to the flow shear term, which represents the importance of natural convection relative to forced convection; and $Prandtl$ ($Pr$) number reflects the ratio of momentum diffusivity to thermal diffusivity, which depends only on the fluid property and state.

**Table 2. Physical Feature Identification for Mixed Convection Case Study**

|  | Physical Feature | Number |
|---|---|---|
| **Inputs** | $\frac{\Delta var_i}{\Delta x_j}$ and $\frac{\Delta^2 var_i}{\Delta x_j \Delta x_i}$ | 10 +15 (2D) |
|  | $Re, Gr, Ri, Pr, R$ | 5 |
| **Outputs** | $\Delta QoI_i = QoI_{HF} - QoI_{LF}$ | 3 (2D) |

\* Variables are $u, v, T, P, k$. QoIs are $u, v, T$.

$$R = \frac{\mu_T}{\mu} = C_\mu \rho \frac{k^2}{\epsilon \mu} \tag{7}$$

$$Re = Re_w{}^\gamma \cdot Re_f{}^{1-\gamma} \tag{8}$$

$$Re_w = \frac{|U| D_{h,w}}{\nu} \tag{9}$$

$$Re_f = \frac{|U| D_{h,f}}{\nu} \tag{10}$$

$$\gamma = \begin{bmatrix} Min\left[Max\left(0, \frac{w}{y} - 1\right), \frac{1}{w}\right], & \text{if cell is adjacent to wall} \\ 0, & \text{if cell is not adjacent to wall} \end{bmatrix} \tag{11}$$

$$Gr = Gr_w{}^\gamma \cdot Gr_f{}^{1-\gamma} \tag{12}$$

$$Gr_w = \frac{g(\rho_w - \rho_f)\rho_f D_{h,w}{}^3}{\mu^2} \tag{13}$$

$$Gr_f = \frac{g(\rho_w - \rho_f)\rho_f D_{h,f}{}^3}{\mu^2} \tag{14}$$

$$Ri = \frac{Gr}{Re^2} \tag{15}$$

$$Pr = \frac{c_p \mu}{\lambda} \tag{16}$$

where $\mu_T$ is turbulent viscosity, $\mu$ is fluid-dynamic viscosity, $k$ is turbulent kinetic energy, $\epsilon$ is turbulent dissipation, $C_\mu$ is a constant set, as 0.09 in GOTHIC, $\rho$ is fluid density, $U$ is local velocity, $\nu$ is fluid kinetic viscosity, $w$ is cell width, $y$ is wall distance from cell center, $D_{h,w}$ and $D_{h,f}$ are, respectively, hydraulic diameters in a near-wall cell and a free cell, and $\lambda$ is thermal conductivity. Considering case study is a two-dimensional mixed convection problem, QoIs in this case are two velocities ($u, v$) and temperature ($T$).

These case studies are applied to (1) evaluate the predictive performance of FSM, (2) explore the importance of training data size and similarity (the similarity is quantified using the mean of KDE distance between training data and target data, expressed as Equation (6)), and (3) investigate the relationship between data similarity and prediction error. The metric of prediction error is normalized root mean-squared errors (NRMSEs) of variables, as calculated in Equation (17).

$$NRMSE_{prediction} = \frac{\sqrt{\frac{1}{n}\Sigma(QoI_{HF,i} - QoI_{predicted,i})^2}}{\frac{1}{n}\Sigma QoI_{HF,i}} \qquad (17)$$

### 4.1. Extrapolation of Geometry (Aspect Ratio)

In this case, three cavities with different aspect ratios are modeled, as shown in Figure 4. The injection condition and geometry parameters are listed in Table 3. Dataset A contains square modeling cases. Dataset B and C respectively contain rectangular modeling cases, with aspect ratios of 1/0.8 and 0.8/1. For each case, one high-fidelity simulation is performed by fine-mesh Star CCM+, four low-fidelity simulations are performed by GOTHIC with different coarse meshes (1/10, 1/15, 1/25, 1/30 m). Each case in A generates 1850 data points, while each case in B and C generates 1480 data points.

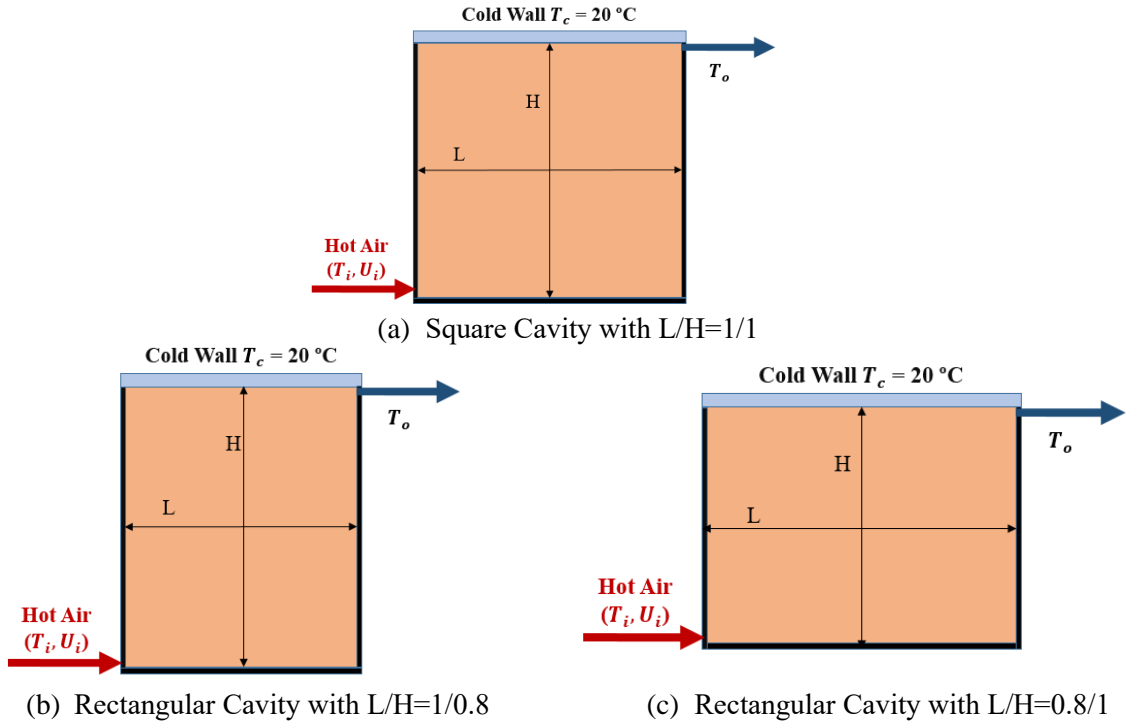

(a) Square Cavity with L/H=1/1

(b) Rectangular Cavity with L/H=1/0.8   (c) Rectangular Cavity with L/H=0.8/1

**Figure 4. Three Mixed Convection Models with Different Aspect Ratios**

Table 3. Geometry and Injection Conditions of Datasets in Extrapolation of Geometry

| Dataset | | Geometry | Aspect Ratio | Injection Temperature (°C) | Injection Rate (m/s) | Data Size |
|---|---|---|---|---|---|---|
| A | 1 | Square | L/H = 1/1 | 30 | 0.1 | 12*1850 |
| | 2 | | | 33 | 0.2 | |
| | 3 | | | 36 | 0.3 | |
| | 4 | | | 39 | 0.4 | |
| | 5 | | | 42 | 0.1 | |
| | 6 | | | 45 | 0.2 | |
| | 7 | | | 48 | 0.3 | |
| | 8 | | | 51 | 0.4 | |
| | 9 | | | 54 | 0.1 | |
| | 10 | | | 57 | 0.2 | |
| | 11 | | | 60 | 0.3 | |
| | 12 | | | 63 | 0.4 | |
| B | | Rectangular | L/H = 1/0.8 | 45–51 | 0.2–0.4 | 3*1480 |
| C | | Rectangular | L/H = 0.8/1 | 45–51 | 0.2–0.4 | 3*1480 |

As shown in Table 4, Tests 1–3 are designed to investigate how the training-data size affects the predictive capability of this data-driven approach. To predict the simulation errors of cases with rectangular cavities, different training-data points are used for the training of DFNN model. By using the dimensionality-reduction technique t-SNE method , the physics-coverage condition of the target case can be qualitatively visualized as shown in Figure 5. The data points of a rectangular case are covered or overlapped by the training data points in square cases, even though globally, testing dataset is an extrapolation of geometry to the training dataset. The PCC of target case is determined as GELI condition. The values of mean of KDE distance are also listed in Table 4. Thus, when the mean of KDE distance decreases, prediction accuracy increases, even if the training-data size decreases. Lower mean of KDE distance represents higher similarity of training data to testing data. It implies that data similarity should be considered in the first place to construct the training database. Adding too much dissimilar or irrelevant data may mislead the training and predictive capability of DFNNs. The comparisons between original low-fidelity simulation results and corrected results based on FSM prediction of Test 3 are shown in Figure 6.

Table 4. Physics Coverage Conditions in Extrapolation of Geometry Case Study

| Test NO. | Training Dataset | Testing Dataset | PCC | NRMSE (u) | NRMSE (v) | NRMSE (T) | Mean of KDE Distance |
|---|---|---|---|---|---|---|---|
| 1 | A (1 ~ 12) | B+C | GELI | 0.2712 | 0.3558 | 0.0223 | 0.3190 |
| 2 | A (4 ~ 10) | B+C | | 0.2640 | 0.3514 | 0.0243 | 0.2894 |
| 3 | A (6 ~ 8) | B+C | | 0.0278 | 0.0287 | 0.0022 | 0.2773 |
| 4 | B+C | A (6 ~ 8) | GILI | 0.0151 | 0.0140 | 0.0009 | 0.2687 |

* Global Physics: Geometry (Aspect Ratio); Local Physics: Physical Feature Group.

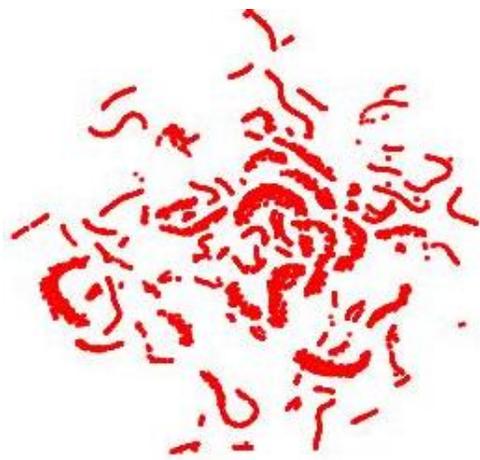 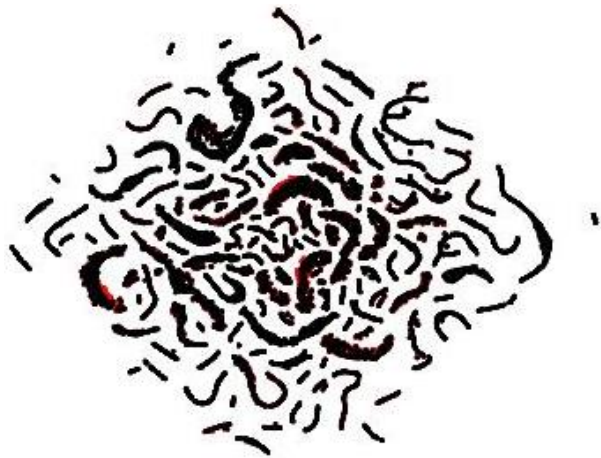

(a) Red: Rectangular Data (B+C)  (b) Test 1: A (1-12) over B+C

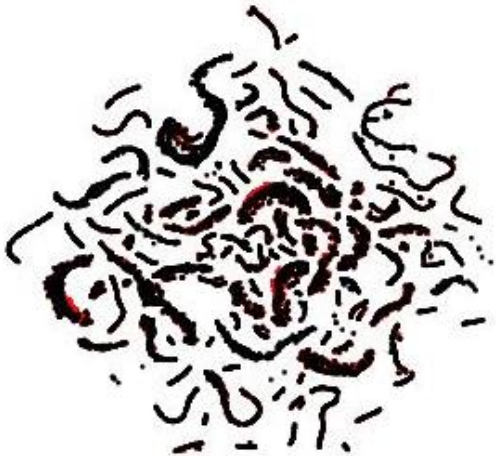 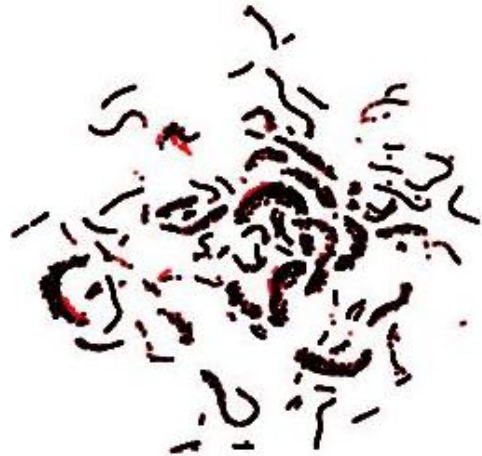

(c) Test 2: A (4-10) over B+C  (d) Test 3: A (6-8) over B+C

**Figure 5. Physical Feature Coverage Conditions between "Rectangular" Cases and "Square" Cases**

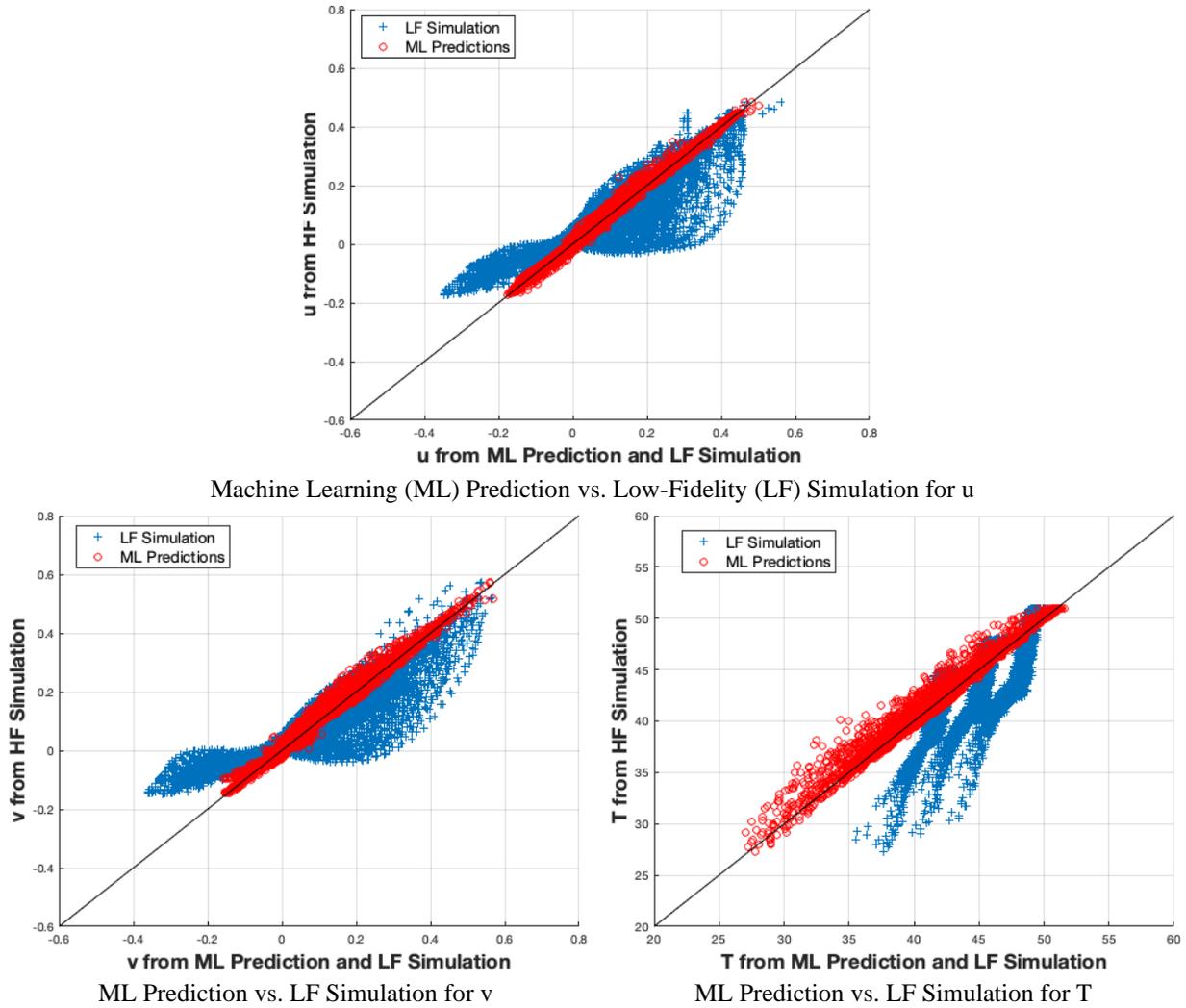

Machine Learning (ML) Prediction vs. Low-Fidelity (LF) Simulation for u

ML Prediction vs. LF Simulation for v

ML Prediction vs. LF Simulation for T

**Figure 6. Comparisons between Original Low-Fidelity Simulation Results and Corrected Results using FSM Approach for Test 3**

Test 4 is identified as a situation in GILI condition because the square cavity can be considered an interpolation of these two rectangular cavities. For data training using the same DFNN structure and initial hyper parameters, the prediction errors (NRMSEs) are listed in Table 3 and are much smaller than the tests in GELI condition. The comparisons between the original low-fidelity simulation results and corrected results based on FSM prediction of Test 4 are shown in Figure 7. The FSM approach has better predictive capability in GILI condition than in GELI condition. The mean of KDE distance of Test 4 (GILI) is smaller than that of the other three tests, which indicates that the training data of Test 4 have a higher data-similarity level than others.

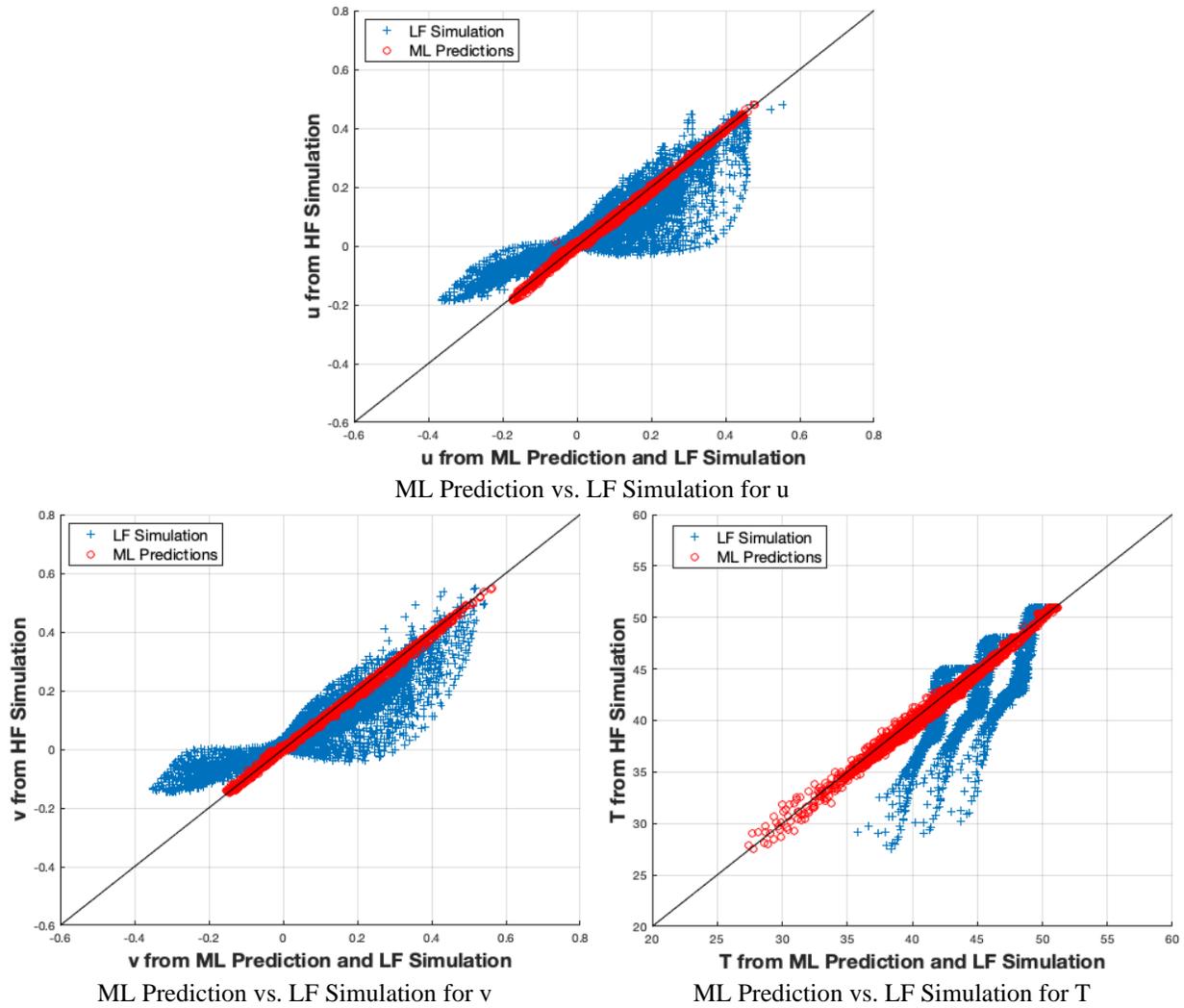

ML Prediction vs. LF Simulation for u

ML Prediction vs. LF Simulation for v     ML Prediction vs. LF Simulation for T

**Figure 7. Comparisons between Original Low-Fidelity Simulation Results and Corrected Results using FSM Approach for Test 4**

### 4.2. Extrapolation of Boundary Condition

In this case, cavities with two different boundary conditions are modeled, as shown in Figure 8. The boundary and injection conditions are listed in Table 5. Dataset A contains cases 1–12 from the previous case study discussed in Section 4.2. Dataset D, E, F and G, respectively, contain the cases with fixed uniform heat flux 100, 120, 150 and 200 W/m$^2$ on top wall.

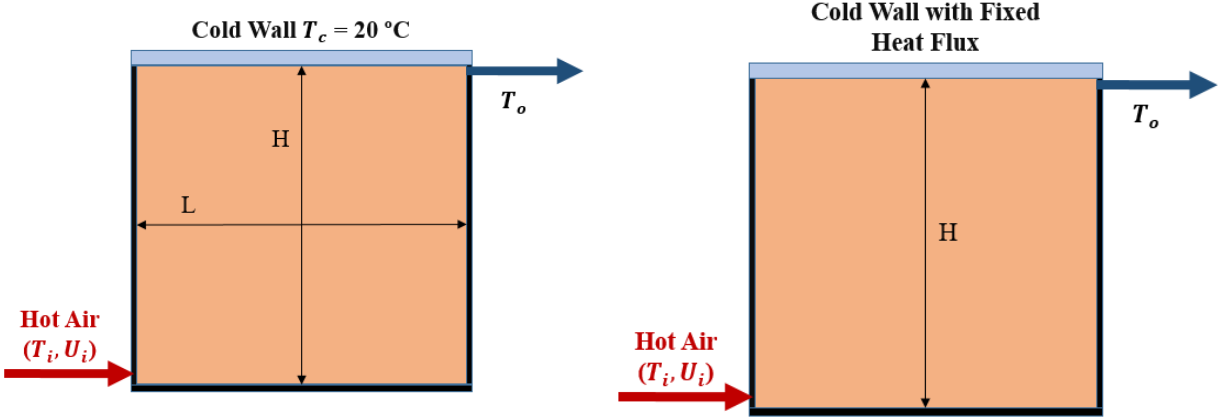

Figure 8. Two Mixed Convection Models with Different Boundary Conditions

Table 5. Boundary and Injection Conditions of Datasets in Extrapolation of Boundary Condition

| Dataset | | Boundary Condition (Averaged Heat Flux from Top Wall) | Injection Temperature (°C) | Injection Rate (m/s) | Data Size |
|---|---|---|---|---|---|
| A | 1 | 26.6 W/m$^2$ | 30 | 0.1 | 12*1850 |
| | 2 | 55.0 W/m$^2$ | 33 | 0.2 | |
| | 3 | 88.7 W/m$^2$ | 36 | 0.3 | |
| | 4 | 128.9 W/m$^2$ | 39 | 0.4 | |
| | 5 | 57.6 W/m$^2$ | 42 | 0.1 | |
| | 6 | 104.6 W/m$^2$ | 45 | 0.2 | |
| | 7 | Cold Top Wall with Fixed T = 20°C — 153.4 W/m$^2$ | 48 | 0.3 | |
| | 8 | 207.8 W/m$^2$ | 51 | 0.4 | |
| | 9 | 87.8 W/m$^2$ | 54 | 0.1 | |
| | 10 | 153.1 W/m$^2$ | 57 | 0.2 | |
| | 11 | 216.7 W/m$^2$ | 60 | 0.3 | |
| | 12 | 284.8 W/m$^2$ | 63 | 0.4 | |
| D | | 100 W/m$^2$ | 48 | 0.3 | 1850 |
| E | | Cold Top Wall with Fixed Heat Flux — 120 W/m$^2$ | | | 1850 |
| F | | 150 W/m$^2$ | | | 1850 |
| G | | 200 W/m$^2$ | | | 1850 |

Tests 5–8 are designed as listed in Table 6. The comparisons between original low-fidelity simulation results, and corrected results, using FSM prediction on Test 6, are displayed in Figure 9. FSM approach presents great predictive capability for velocities, some predictions of temperature errors are not good enough to correct the original low-fidelity simulation results.

**Table 6. Physics Coverage Conditions in Extrapolation of Boundary Condition Case Study**

| Test NO. | Training Dataset | Testing Dataset | PCC | NRMSE (u) | NRMSE (v) | NRMSE (T) | Mean of KDE Distance |
|---|---|---|---|---|---|---|---|
| 5 | A | D | GELI (Partially GELE) | 0.254 | 0.297 | 0.034 | 0.2466 |
| 6 | | E | | 0.156 | 0.202 | 0.026 | 0.2444 |
| 7 | | F | | 0.280 | 0.305 | 0.043 | 0.2493 |
| 8 | | G | | 0.623 | 0.658 | 0.087 | 0.2566 |

* Global Physics: Boundary Condition; Local Physics: Physical Feature Group.

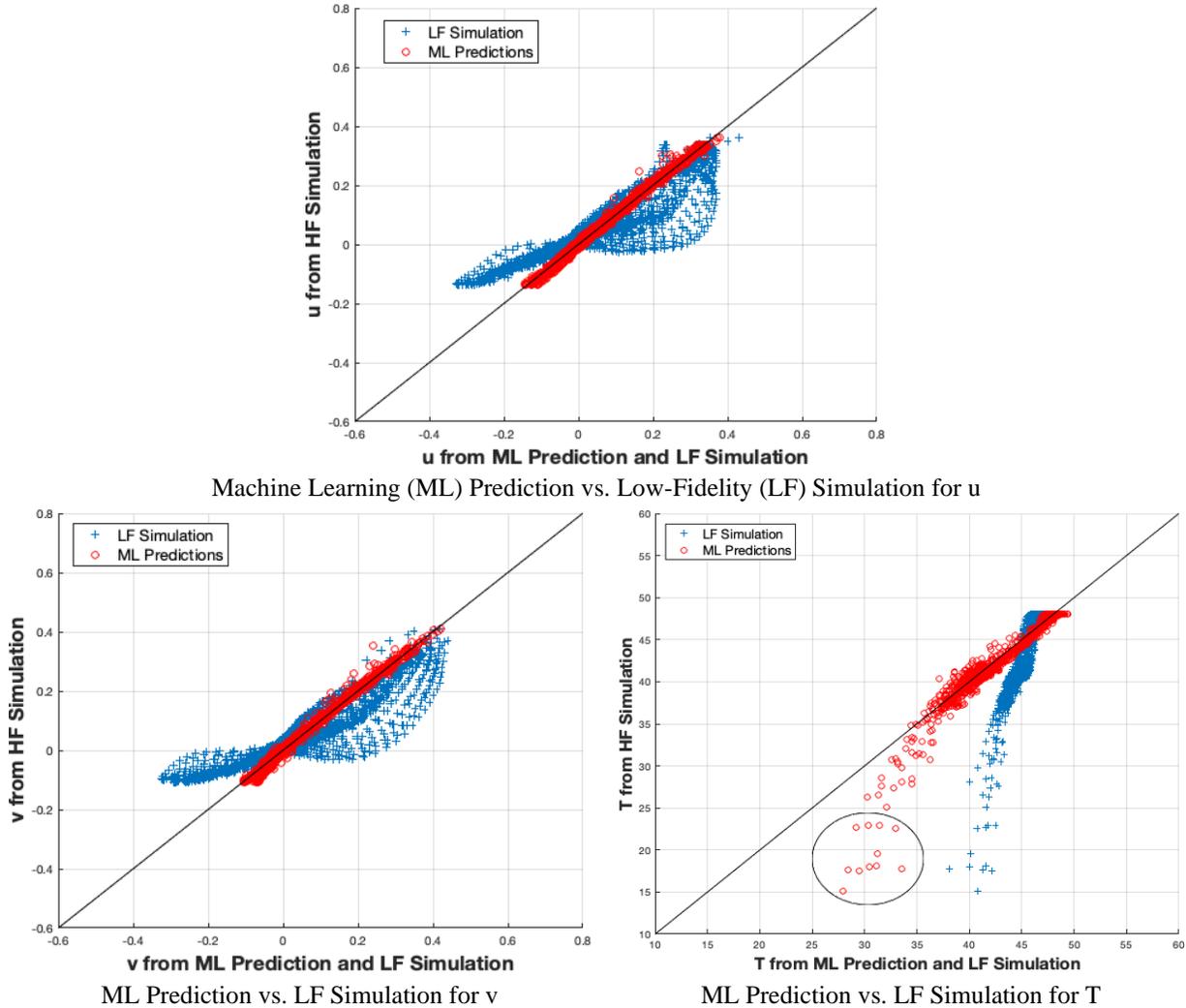

Machine Learning (ML) Prediction vs. Low-Fidelity (LF) Simulation for u

ML Prediction vs. LF Simulation for v

ML Prediction vs. LF Simulation for T

**Figure 9. Comparisons between Original Low-Fidelity Simulation Results and Corrected Results using FSM Approach for Test 6**

To determine the locations of these "bad" predictions, the distributions of DFNN prediction error of temperature for four mesh configurations are plotted in Figure 10. These points, mainly located at the top-left part of the cavity no matter which mesh configuration is applied, have been marked in red circles. In this area, heat transfer is underestimated in both low-fidelity simulation and machine-learning prediction. The simulations of training cases with fixed temperatures should have different heat fluxes from different cells to the cold wall because of the locations of the injection and vent. The cells at top-right part have higher heat fluxes than the cells at top-left part when convection reaches a steady state. For example, in Case A-7, although the averaged heat flux is 153.4 W/m$^2$, the real heat flux at the top-left part is much lower than 153.4 W/m$^2$. The heat flux is not uniform along the top wall. The temperature difference between the cell at the left corner and the cell at top-right corner for Case A-7 is about 8°C in low-fidelity simulations and 18°C in high-fidelity simulations. Here, Case A-7 is used to compare with testing cases because they have the same injection rate and temperature.

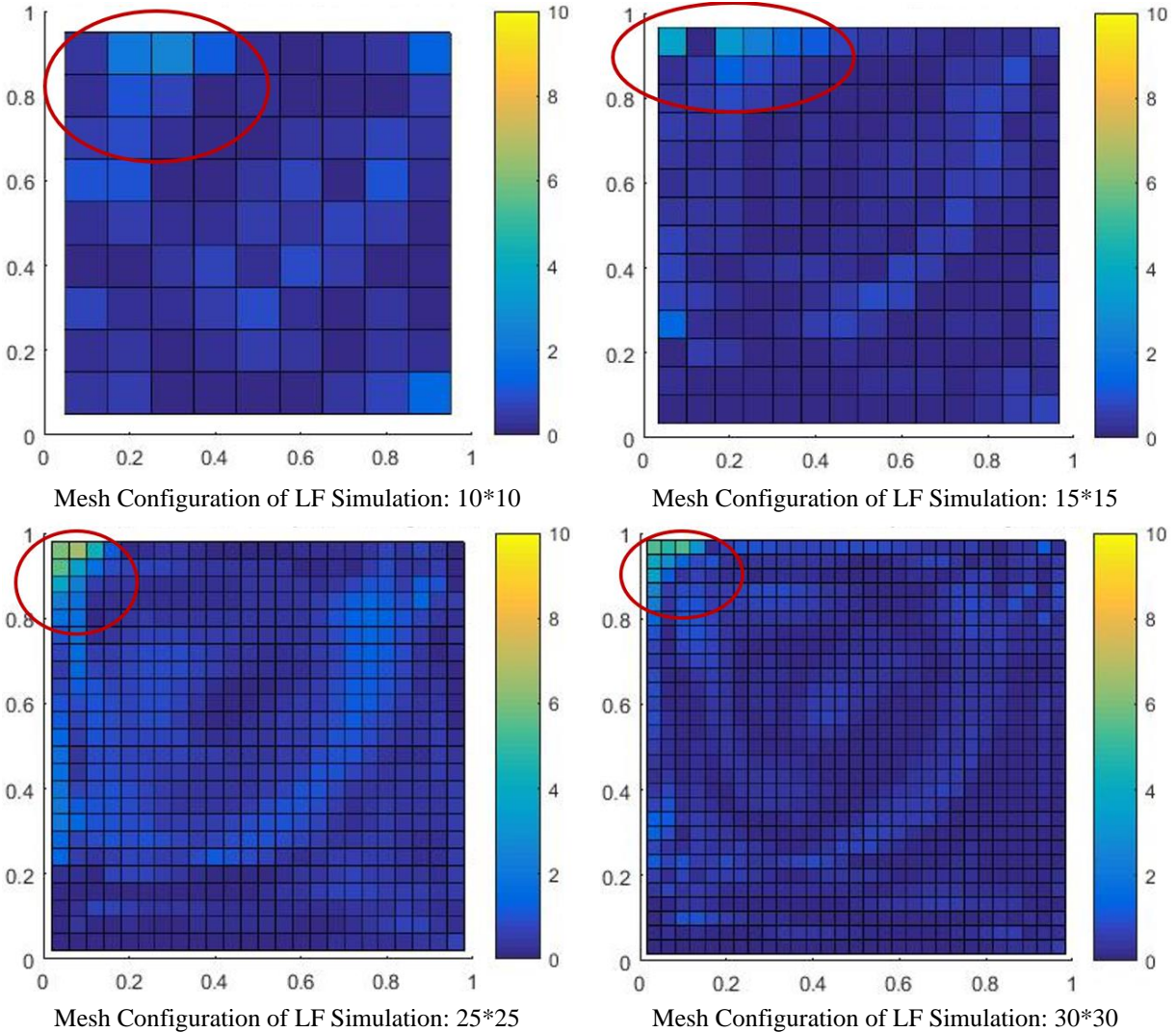

Mesh Configuration of LF Simulation: 10*10    Mesh Configuration of LF Simulation: 15*15

Mesh Configuration of LF Simulation: 25*25    Mesh Configuration of LF Simulation: 30*30

**Figure 10. Distribution of DFNN Prediction Errors of Temperature for Different Mesh Configurations in Test 6**

In simulations of the testing cases, fixed heat fluxes are enforced for all top cells, which requires a "stronger" heat-transfer capability at the left part than in the training cases. According to high-fidelity simulation results, it leads to a large temperature difference (30°C for Case E) between the cell at the left corner and the cell at the top-right corner. However, the responding low-fidelity simulation only has a 10°C temperature difference. The reason is that the convective correlations selected in low-fidelity simulation do not have the necessary, significant heat-removal capability. The heat-transfer coefficients calculated are much lower than the ones calculated in high-fidelity simulation. To satisfy the fixed heat flux, the temperatures in top cells must be high enough to keep large temperature differences between the fluid and the top cold wall.

However, these convective correlations perform better for training cases in which the strong heat-transfer capability is not needed. This underlying physics in testing cases is not learned by the well-trained DFNNs. The DFNNs estimate relatively high temperatures in top cells for the testing case in which the temperature calculated in high-fidelity simulation is lower because the heat-transfer capability is more significant than the DFNNs expect. This sort of "wrong" learning in the top-left part also reflects in the predictions for Dataset D, F, and G. Therefore, the PCC of these testing cases can be determined as GELI for the major part of testing data, but GELE for the data at the top-left part. The NRMSE of predictions are compared in Table 6. Smaller KDE distance implies better predictions.

### 4.3. Extrapolation of Dimension

In this case, cavities with different dimensions are modeled. The boundary and injection conditions are listed in Table 7. Dataset H, I, and J, respectively, contain the cases with length equal to 1.2, 1.5 and 2 m. Dataset A is applied as training data, and datasets H, I, and J are set as testing cases. The same mesh configuration of 30 × 30 is applied to all low-fidelity simulations for testing cases.

Table 7. Description of Datasets in Extrapolation of Dimension Case Study

| Dataset | | Dimension | Mesh Size | Injection Temperature (°C) | Injection Rate (m/s) | Data Size |
|---|---|---|---|---|---|---|
| A | 1~12 | 1m × 1m | 1/10, 1/15, 1/25, 1/30 m | 30–63 | 0.1–0.4 | 12*1850 |
| H | | 1.2m × 1.2m | 1/25 m | | | 900 |
| I | | 1.5m × 1.5m | 1/20 m | 48 | 0.3 | 900 |
| J | | 2m × 2m | 1/15 m | | | 900 |

Three tests are designed as listed in Table 8, with the NRMSEs of predictions. It shows that the smaller mean of KDE distance implies smaller NRMSEs and better predictions. Physical-feature similarity of tests is compared in Figure 11, where red points represent testing data, and black points represent training data. For Test 9, part of the testing data is covered by training data. For Tests 10 and 11, testing data are rarely covered, which can be considered a GELE condition. Both the global (dimension) and local (physical features) physics of testing cases are extrapolative. This shows that the FSM approach does not present good predictive capability in GELE condition. The mesh sizes applied in testing cases are much smaller than the ones in training cases, which greatly affects the similarity of physical features.

Table 8. Prediction Results of Extrapolation of Dimension Case Study

| Test NO. | Training Dataset | Testing Dataset | PCC | NRMSE (u) | NRMSE (v) | NRMSE (T) | Mean of KDE Distance |
|---|---|---|---|---|---|---|---|
| 9 | | H (1.2 m) | GELI | 0.291 | 0.378 | 0.043 | 0.2539 |
| 10 | A | I (1.5 m) | + | 0.552 | 0.785 | 0.083 | 0.2819 |
| 11 | | J (2 m) | GELE | 0.803 | 1.163 | 0.123 | 0.3059 |

* Global Physics: Dimension; Local Physics: Physical Feature Group.

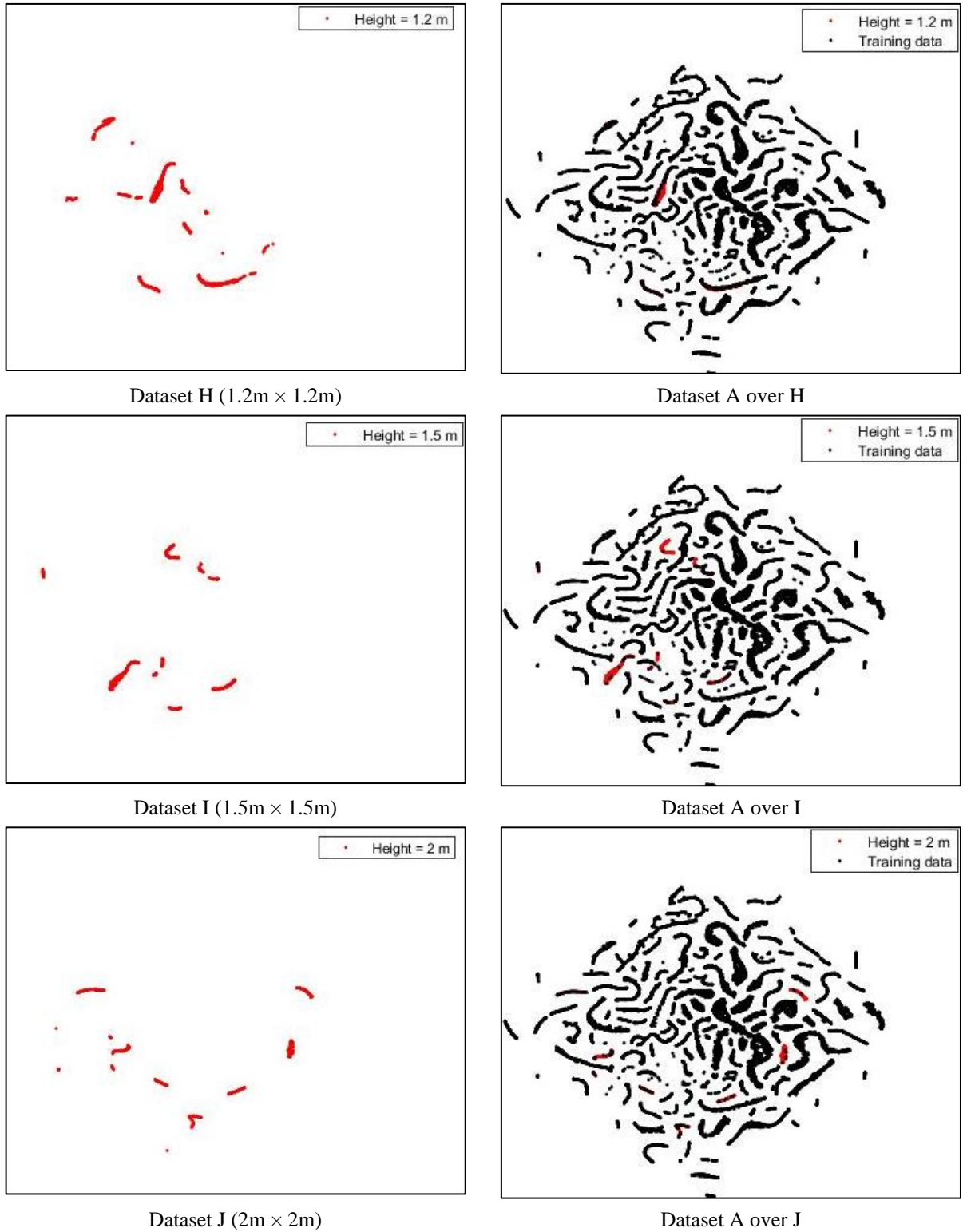

Dataset H (1.2m × 1.2m)  Dataset A over H

Dataset I (1.5m × 1.5m)  Dataset A over I

Dataset J (2m × 2m)  Dataset A over J

**Figure 11. Physical Feature Similarity in Extrapolation of Dimension Case Study**

To improve the predictive performance of FSM, the similarity between training and testing data should be enhanced. The current low similarity between training and testing data owes to large differences between global dimension and local mesh size. The impact of mesh size was considered in the identification of local physical features because a major part of physical features is defined with local mesh size as a key parameter. Therefore, one way to improve the similarity between training and testing data is to consider the effect of global conditions in the definitions of local physical features. One part of local physical features is the gradients of variables; here in the new test, all of these gradients are nondimensionalized based on global conditions. Table 9 lists the differences of definitions between previous physical features and new ones. $u_{inj}$ and $T_{inj}$ represent injection rate and temperature, $P_{ini}$ is initial pressure in the cavity.

**Table 9. Some Examples of New Physical Features Nondimensionalized based on Global Conditions**

| Physical Feature | 1-order Gradients of Variables | | | | | 2-order Gradients of Variables | | |
|---|---|---|---|---|---|---|---|---|
| Previous | $\dfrac{\Delta u}{\Delta x}$ | $\dfrac{\Delta v}{\Delta y}$ | $\dfrac{\Delta T}{\Delta y}$ | $\dfrac{\Delta P}{\Delta x}$ | $\dfrac{\Delta k}{\Delta x}$ | $\dfrac{\Delta^2 u}{\Delta x \Delta y}$ | $\dfrac{\Delta^2 T}{\Delta x^2}$ | $\dfrac{\Delta^2 k}{\Delta y^2}$ |
| New | $\dfrac{\frac{\Delta u}{\Delta x}}{\frac{u_{inj}}{width}}$ | $\dfrac{\frac{\Delta v}{\Delta y}}{\frac{u_{inj}}{height}}$ | $\dfrac{\frac{\Delta T}{\Delta y}}{\frac{T_{inj}}{height}}$ | $\dfrac{\frac{\Delta P}{\Delta x}}{\frac{P_{ini}}{width}}$ | $\dfrac{\frac{\Delta k}{\Delta x}}{\frac{u_{inj}^2}{width}}$ | $\dfrac{\frac{\Delta^2 u}{\Delta x \Delta y}}{\frac{u_{inj}^2}{width \cdot height}}$ | $\dfrac{\frac{\Delta^2 T}{\Delta x^2}}{\frac{T_{inj}}{width^2}}$ | $\dfrac{\frac{\Delta^2 k}{\Delta y^2}}{\frac{u_{inj}^4}{height^2}}$ |

Table 10 displays FSM prediction errors with the previous and new physical features and the original low-fidelity simulation errors for Tests 9, 10, and 11. With new physical features considering global condition information, the DFNNs provide more-accurate predictions on simulation errors to improve the low-fidelity simulation results. Figure 12 compares the original low-fidelity simulation results, FSM predictions with previous and new physical features for Test 9. The improvement in temperature predictions is significant.

**Table 10. Prediction Results of Extrapolation of Dimension Case Study with New Physical Features**

| Test NO. | Training Dataset | Description | NRMSE (u) | NRMSE (v) | NRMSE (T) | Mean of KDE Distance |
|---|---|---|---|---|---|---|
| 9 H (1.2 m) | A | FSM (previous PFs) | 0.291 | 0.378 | 0.043 | 0.2039 |
| | | **FSM (new PFs)** | **0.231** | **0.261** | **0.015** | **0.2059** |
| | | LF Simulation | 1.302 | 1.552 | 0.094 | - |
| 10 I (1.5 m) | | FSM (previous PFs) | 0.552 | 0.785 | 0.083 | 0.2819 |
| | | **FSM (new PFs)** | **0.410** | **0.508** | **0.032** | **0.2218** |
| | | LF Simulation | 1.214 | 1.444 | 0.106 | - |
| 11 J (2 m) | | FSM (previous PFs) | 0.803 | 1.163 | 0.123 | 0.3059 |
| | | **FSM (new PFs)** | **0.595** | **0.766** | **0.065** | **0.2507** |
| | | LF Simulation | 1.044 | 1.232 | 0.118 | - |

\* Global Physics: Dimension; Local Physics: Physical Feature Group.

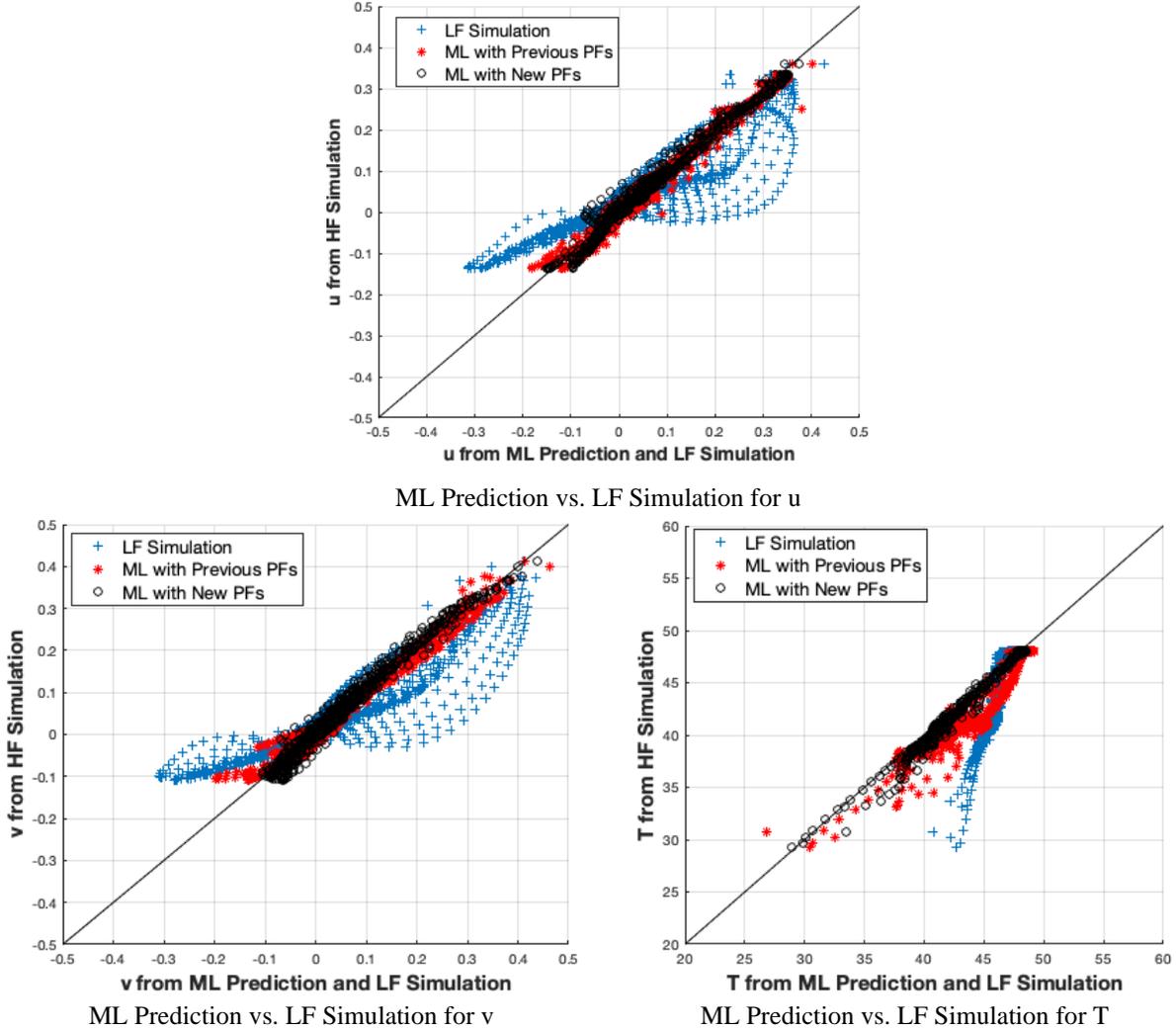

ML Prediction vs. LF Simulation for u

ML Prediction vs. LF Simulation for v    ML Prediction vs. LF Simulation for T

**Figure 12. Comparisons among Original Low-Fidelity Simulation Results, FSM Predictions with Previous Physical Features and with New Physical Features for Test 9**

## 5. CONCLUSIONS

In this paper, a data-driven approach, feature-similarity measurement, is proposed to identify the local physical features, measure their similarity, and investigate the relationship between physical-feature similarity and machine-learning prediction accuracy in the global extrapolation through local interpolation condition. In GELI condition, the similarity of local physics and patterns can be used to bridge the global scale gap. FSM treats model error, mesh-induced error, and scaling uncertainty together, and estimates the simulation error by exploring local similarity in multiscale data using deep learning. The local patterns are represented by a set of physical features, which integrate the information from the physical system of interest, empirical correlations and the effect of mesh size. Data similarity is quantitatively measured using KDE distance and qualitatively visualized by the t-SNE technique. Case studies show that FSM has good predictive capability in GELI and GILI conditions. Although FSM only has limited predictive capability in GELE condition, some methods can be applied to take global condition information into the definitions of local physical features and enhance data similarity between training and target cases. Prediction accuracy increases with increase of data similarity of local physical features.

**ACKNOWLEDGMENTS**

Authors gratefully acknowledge the support by the Idaho National Laboratory's National University Consortium (NUC) program and the INL Laboratory Directed Research & Development (LDRD) program under DOE Idaho Operations Office Contract DE-AC07-05ID14517. The authors are also grateful for partial support from the U.S. Department of Energy's Consolidated Innovative Nuclear Research program via the Integrated Research Project on "Development and Application of a Data-Driven Methodology for Validation of Risk-Informed Safety Margin Characterization Models" under the grant DE-NE0008530. GOTHIC incorporates technology developed for the electric power industry under the sponsorship of EPRI, the Electric Power Research Institute. This work was completed using a GOTHIC license for educational purposes provided by Zachry Nuclear Engineering, Inc.